\documentclass[10pt,twocolumn,letterpaper]{article}

\usepackage{cvpr}
\usepackage{times}
\usepackage{epsfig}
\usepackage{graphicx}
\usepackage{amsmath}
\usepackage{amssymb}

\usepackage[breaklinks=true,bookmarks=false]{hyperref}

\cvprfinalcopy 

\def\cvprPaperID{****} 

\usepackage{enumitem}
\usepackage{booktabs}
\usepackage{wrapfig}
\usepackage{subfig}
\newcommand{\minisection}[1]{\vspace{3mm}\noindent{\bf #1}}
\makeatletter
\def\blfootnote{\xdef\@thefnmark{}\@footnotetext}
\makeatother

\begin{document}
\title{Facial Landmark Detection with Tweaked Convolutional Neural Networks}

\author{
\begin{tabular}{c@{\extracolsep{0.6cm}}c@{\extracolsep{0.6cm}}c@{\extracolsep{0.6cm}}c@{\extracolsep{0.6cm}}c}
Yue Wu$^{1}$* & Tal Hassner$^{1,2}$* & KangGeon Kim$^3$ & G\'{e}rard Medioni$^3$ & Prem Natarajan$^1$\\
\end{tabular}\\
{\small $^{1}$ Information Sciences Institute, USC, CA, USA}\\
{\small $^{2}$ The Open University of Israel, Israel}\\
{\small $^{3}$ Institute for Robotics and Intelligent Systems, USC, CA, USA}\\
}

\maketitle


\begin{abstract}
We present a novel convolutional neural network (CNN) design for facial landmark coordinate regression. We examine the intermediate features of a standard CNN trained for landmark detection and show that features extracted from later, more specialized layers capture rough landmark locations. This provides a natural means of applying differential treatment midway through the network, tweaking processing based on facial alignment. The resulting {\em Tweaked CNN model} (TCNN) harnesses the robustness of CNNs for landmark detection, in an appearance-sensitive manner without training multi-part or multi-scale models. Our results on the AFLW, AFW, and 300W benchmarks show improvements over existing work. We further provide results on the Janus benchmark, demonstrating the benefit of our better alignment in face verification.\blfootnote{* Denotes joint first authorship / equal contribution.} 
\end{abstract}

\section{Introduction}

Recent years brought increasing interest in facial landmark detection techniques. To great extent, this is triggered by the many applications facial landmarks have in head pose estimation~\cite{demirkus2015hierarchical,zhu2012face}, emotion classification~\cite{ding2013facial,levi2015emotion}, face alignment in 2D~\cite{cao2014face,eidinger2013age,xiong2013supervised} and~3D (e.g., {\em frontalization}~\cite{hassner2015effective}) and, of course, face recognition (see, e.g.,~\cite{sun2014deep} and many others).

This task is particularly daunting considering the real-world, unconstrained imaging conditions typically assumed: Images often portray faces in myriads of poses, expressions, occlusions and more, any one of which can affect landmark appearances, locations or even presence. 
\begin{figure}[t!]
\begin{center}
\begin{tabular}{cc}
\rotatebox{90}{\hspace{30mm}{\em RGB}} & \includegraphics[width=.75\linewidth,clip,trim = 3mm 3mm 3mm 3mm]{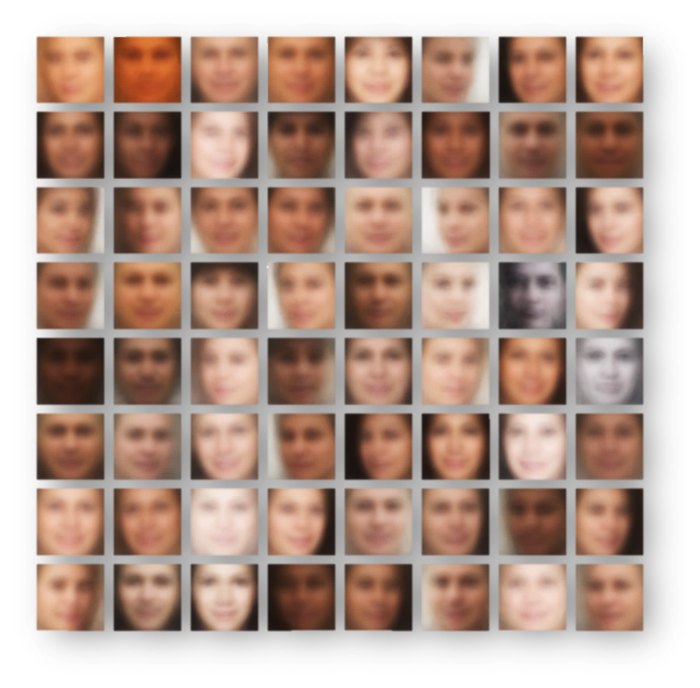}\\
\rotatebox{90}{\hspace{12mm}{\em Deep intermediate features}} & \includegraphics[width=.75\linewidth,clip,trim = 3mm 3mm 3mm 3mm]{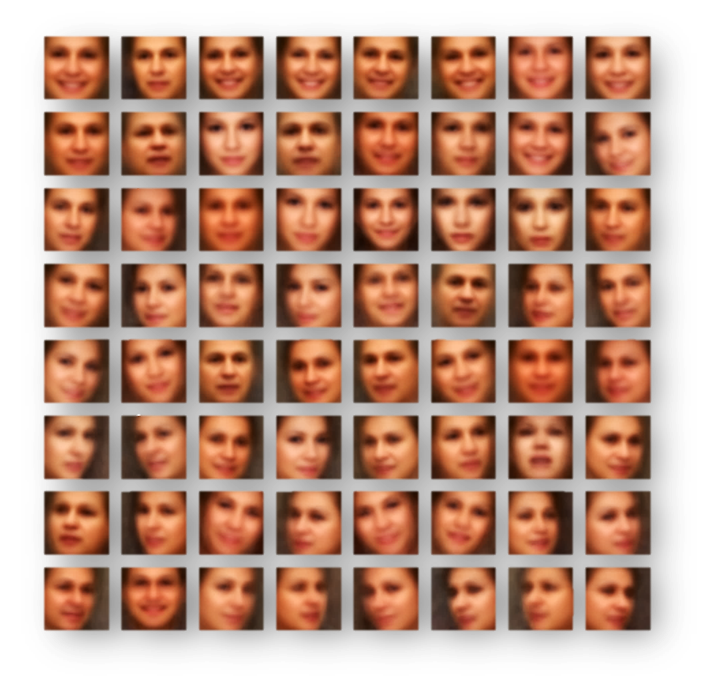}
\end{tabular}
\vspace{-3mm}
\end{center}
\caption{{\em Average images for 64 face clusters.} Top: clusters computed using RBG values. These appear misaligned (blurry) and strongly influenced by intensities. Bottom: Images clustered using features from an intermediate layer of a network trained to regress facial landmarks. These are clearly better aligned. We leverage on this to tweak network processing based on intermediate representations. (Note: Both results produced using the same images.)} \label{fig:clusters}
\vspace{-3mm}
\end{figure}

Many effective methods were proposed to handle these challenges. Several use classifiers and robust representations to search for specific facial parts, further disambiguating detections by constraining landmark arrangements~\cite{zhu2012face,baltrusaitis2013constrained,yu2013pose}. Others regress detections directly~\cite{cao2014face,xiong2013supervised,burgos2013robust,ghiasi2014occlusion,kazemi2014one,ren2014face,wu2015robust}. These methods are known to be limited in the pose variations they can account for, leading some to estimate pose prior to detection~\cite{yang2015face}. Some are further motivated by specific appearance variations (e.g., pose~\cite{zhu2012face,yu2013pose} or occlusions~\cite{burgos2013robust,ghiasi2014occlusion}); hence, their performance may not carry over to appearance variations unaccounted for in their design. 

In line with the recent success of deep learning methods, a number of such techniques were proposed for landmark coordinate regression~\cite{liang2015unconstrained,sun2013deep,zhang2014coarse,zhang2014facial,zhang2015learning}. These methods regress landmark positions directly from image intensities, naturally learning landmark appearance and location variations from huge training sets. To do so, they learn multiple part models~\cite{liang2015unconstrained,sun2013deep}, hierarchical representations~\cite{liang2015unconstrained,zhang2014coarse} or infuse networks with additional labeled attribute data~\cite{zhang2014facial,zhang2015learning}. 



We present a novel convolutional neural network (CNN) for face landmark regression. Contrary to others, our design {\em does not involve multiple part models}, it is {\em naturally hierarchical} and requires {\em no auxiliary labels} beyond landmarks. 

We are motivated by recent reports that features from intermediate layers of deep networks become progressively task-specific at deeper layers~\cite{Aubry15,yosinski2014transferable}. As we later show, when trained to detect facial landmarks these specialized features actually reflect alignment quite well. Fig.~\ref{fig:clusters} illustrates this, showing 64 cluster centers -- averages of cluster images -- comparing clusters computed using input RGB values (top) vs. input features from the first dense layer of face landmark regression CNN (Sec.~\ref{sec:vanilla}). RGB clusters clearly reflect image intensities. By comparison, intermediate feature clusters appear to contain well-aligned faces with similar poses, apparent by their sharp average faces. 

We leverage on the fact that deep intermediate features naturally capture alignment and introduce our {\em Tweaked CNN} (TCNN). TCNN automatically diverts deep features to separate, specialized ({\em tweaked}) processing. Tweaking implies fine-tuning the final layers for particular head poses. Thus, heavy processing at the early, convolutive layers is shared amongst all images; differential, specialized treatment is applied only in the final layers. We explain how overfitting is avoided when fine-tuning the final layers, despite limited training data. Finally, we show TCNN to outperform existing state of the art, with an efficient architecture and fewer labels required for training.

\minisection{Contributions.} (1) A first analysis of representations produced at intermediate layers of a deep CNN trained for landmark detection, showing them to be surprisingly good at representing different head poses and (some) facial attributes (Sec.~\ref{sec:analyze}). (2) Our TCNN model which improves CNN-based landmark detection by differently tweaking the processing of different intermediate features (Sec.~\ref{sec:finetune}). To our knowledge, we are the first to use intermediate features in this way. (3) A novel data augmentation method which inflates available training data for fine-tuning on different head poses (Sec.~\ref{sec:augment}). The benefits of these are demonstrated by reporting facial landmark detection accuracies on the AFLW, AFW, and 300W benchmarks and face verification results on Janus CS2.

\section{A CNN for facial landmark detection}\label{sec:vanillaoverview}
We begin by studying a {\em vanilla CNN} trained to detect facial landmarks. Our goal is to ``pop the hood'' off the CNN to better understand its internal representations and the nature of the information they encode. This later guides us in adjusting processing midway through the network.

\begin{figure*}[t!]
	\centering{
	\begin{tabular}{cc}
		\includegraphics[width=.48\linewidth,clip,trim=40mm 60mm 40mm 60mm]{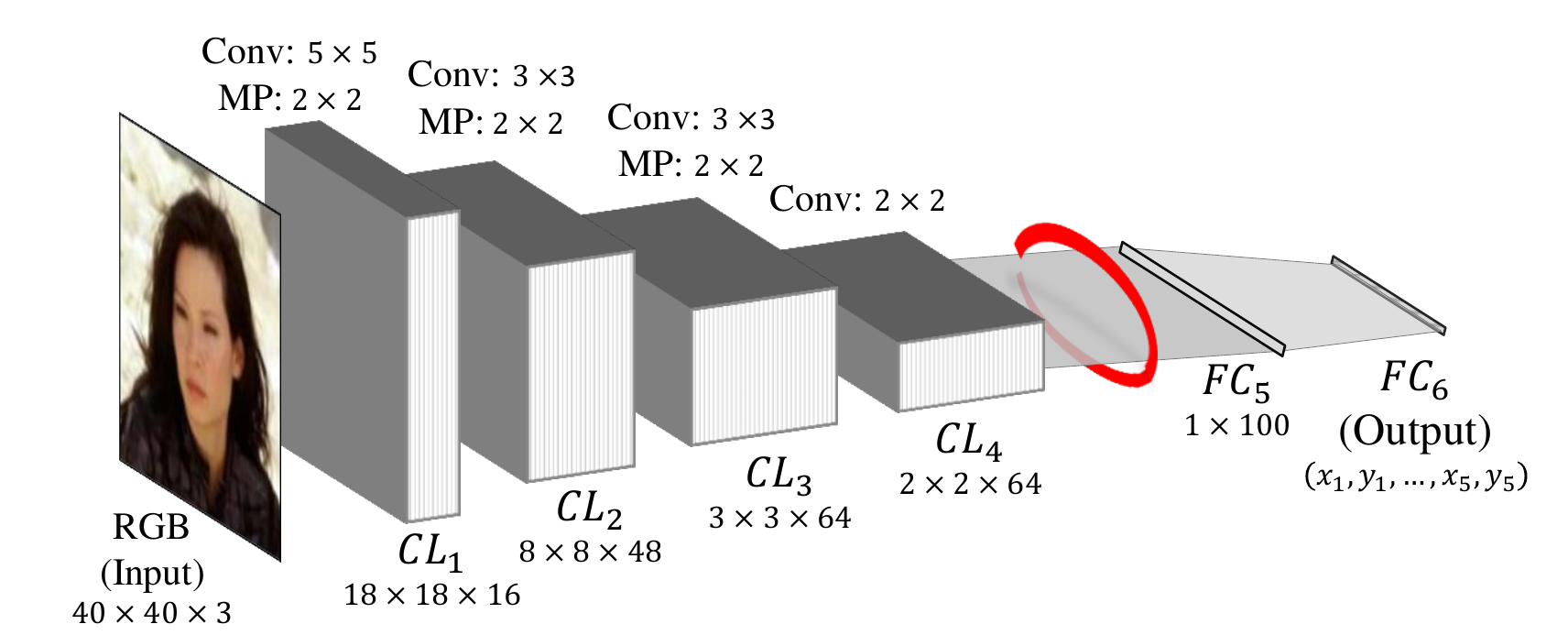} &
		\includegraphics[width=.48\linewidth,clip,trim=40mm 60mm 40mm 60mm]{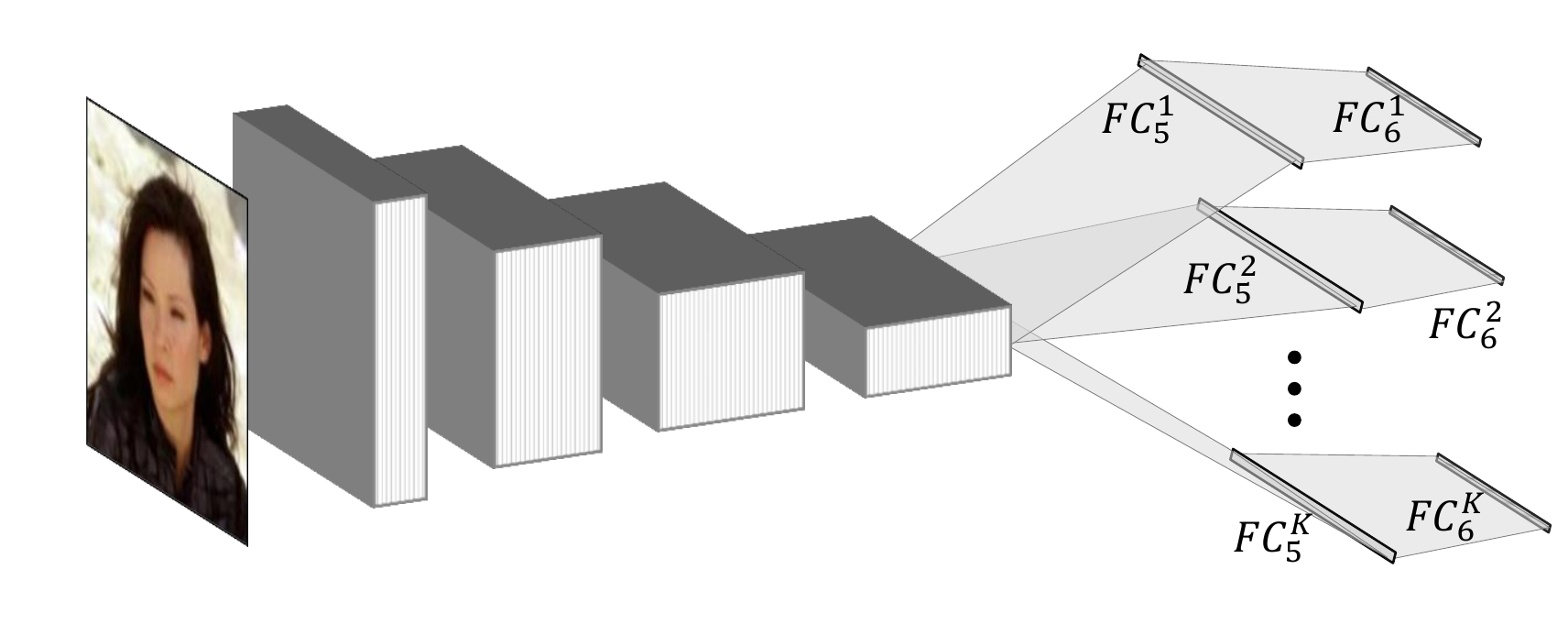}		
		\vspace{-3mm}
	\end{tabular}
	}
	\caption{{\em CNN architectures.} Left: The vanilla network described in Sec.~\ref{sec:vanilla} for facial landmark regression. We show that representations extracted from the input to $FC_5$ (marked in red) are highly specialized and reflect facial alignment. Right: Our Tweaked CNN design, diverting intermediate features to $K$ different subsequent, fine-tuned processes in the same dimensions as the original layers.}
	\label{fig:system}
	\vspace{-3mm}
\end{figure*}

\subsection{Vanilla network design}\label{sec:vanilla}
Our vanilla CNN is loosely based on a state of the art network for facial landmark coordinate regression~\cite{zhang2014facial}. This model was selected to allow direct comparison with previous work as well as to emphasize the contributions of our own TCNN model. Its design is illustrated in Fig.~\ref{fig:system} (left)\footnote{Trained networks and code available from the project page at~\url{www.openu.ac.il/home/hassner/projects/tcnn_landmarks}.}. 

Images are processed by a face detector~\cite{king2009dlib}, returning bounding box coordinates for each face. Bounding boxes are scaled to $40\times 40$ pixels and represented using RGB values, normalized by subtracting the training set mean image and dividing by its standard deviation. 

The network consists of four convolutional layers (denoted $CL_1\hdots CL_4$) with intermittent max pooling layers ({\em stride=$2$}). These are followed by a fully connected (dense) layer, $FC_5$, which is then fully connected to an output with $2\times m$ values for the $m$ landmark coordinates: $\mathbf{P}=(\mathbf{p}_1,\hdots,\mathbf{p}_m)=(x_1,y_1,\hdots,x_m,y_m)$. In our experiments, networks are trained to detect $m=5$ landmarks in the bounding box coordinate frames. 

Absolute hyperbolic tangent is used as an activation function. Finally, unlike~\cite{zhang2014facial}, we use L2, normalized by the inter-ocular distance as the network loss:
\begin{equation}
{\cal{L}}(\mathbf{P}_{i},\hat{\mathbf{P}}_{i})= \frac{ \| \mathbf{P}_{i} - \hat{\mathbf{P}}_{i} \|_2^2 }{ \|{\hat{\mathbf{p}}}_{i,1}-{\hat{\mathbf{p}}}_{i,2}\|_2^2 },\label{eq:loss}
\end{equation}
\noindent where $\mathbf{P}_{i}$ is the $2\times m$ vector of predicted coordinates for training image $I_i$, $\hat{\mathbf{P}}_{i}$ their ground truth locations, and ${\hat{\mathbf{p}}}_{i,1}, {\hat{\mathbf{p}}}_{i,2}$ the reference eye positions. We chose this loss to reflect the standard measure for detection accuracy (Sec.~\ref{sec:results}).

Training used {\em Adam} optimization~\cite{kingma2014adam} with the same images and landmarks from~\cite{sun2013deep,zhang2014facial}. This set contains 5,590 images from the LFW collection~\cite{LFWTech} and 7,876 images from throughout the web, all labeled for five facial landmarks. Whenever the face detector of~\cite{king2009dlib} failed to locate a face, we discarded that image from the training/validation set, further counting such test images as failures when reporting performance in Sec.~\ref{sec:results}. Remaining faces were randomly partitioned, taking 90\% (7,571 images) for training and $1,972$ for validation. As we later report, this network performs comparably with the state of the art, despite its straightforward design, possibly due our use of a loss function more suitable to facial landmark detection.

\subsection{What is the network learning?}\label{sec:analyze}
Once trained, we use the vanilla CNN to extract representations from the input of each layer (including the first: the input RGB values.) Analyzing these, we seek answers to the following questions:

\begin{enumerate}[label={\bf Q\arabic*}]
\item Do similar features reflect similar facial properties?
\item If so, what are these properties?
\item Finally, when (at what layer) are they captured?
\end{enumerate}

A positive answer to {\bf Q1} suggests that prior to landmark detection, the network aggregates particular facial attributes together. This calls to question the effort to artificially introduce this information by other systems (e.g., pose and facial attributes in~\cite{yang2015face,zhang2014facial,zhang2015learning}). Answering {\bf Q2} would tell us what attributes influence network processing. These attributes may provide a means of improving our system's performance. Finally, {\bf Q3} seeks the layer at which the network naturally represents these appearances. At this layer we can assume some knowledge of the attributes of the face in the input image and tweak the remaining stages of our processing accordingly.

If similar representations capture similar properties, then we expect to see this reflected in clusters of these representations. We thus proceed by standard, unsupervised clustering of the representations from each layer.

\minisection{Clustering intermediate features.} Let ${\mathbf f}_{i,l} = f(I_i, L_l)$, and denote the feature vector extracted from the input to layer $L_l$ of our vanilla CNN for training image $I_i$ as $L_{l}\in\{CL_1, \hdots, CL_4, FC_5\}$. We partition the set $\{{\mathbf f}_{i,l}|i=1,\hdots,n\}_{l\in 1\hdots 5}$ into $K$ clusters, $C_{l,k\in 1,\hdots,K}$, using EM to compute Gaussian Mixture Models (GMM)~\cite{bishop2006pattern} and L2 as the feature dissimilarity (corresponding with the normalized L2 used in our cost function, Eq.~\ref{eq:loss}). The vector ${\mathbf f}_{i,l}$ is associated with the cluster with the highest posterior probability:
\begin{equation}
{\mathbf f}_{i,l}\in C_{l,k} ~~ \text{iff} ~~ k=\arg \max_k p( C_{l,k} | {\mathbf f}_{i,l})\label{eq:gmm}
\end{equation}
\noindent This analysis uses $K=64$ clusters per layer. Some per-layer statistics for these clusters are reported in Table~\ref{tab:clusterstats}. Note that GMM provided better results than k-means and was therefore used in all our tests.

\begin{table}[!t]
\setlength{\tabcolsep}{.3em}
\scriptsize{
\caption{{\em Cluster statistics.} Reports median($S_l$) $\pm$ SD($S_l$), where $S_l=\{ |C_{l,k}| | k = 1..64 \}$ is the set of the numbers of images in the clusters of layer $l$.}\label{tab:clusterstats}
\begin{center}
\begin{tabular}{ccccc}
\toprule
$CL_1$ & $CL_2$ & $CL_3$ & $CL_4$ & $FC_5$\\ 
92.0 $\pm$ 98.3 & 110.5 $\pm$ 55.1 & 116.0 $\pm$ 70.0 & 118.5 $\pm$ 81.6 & 110.0 $\pm$ 58.3 \\
\bottomrule
\end{tabular}
\end{center}
}
\vspace{-7mm}
\end{table}

\minisection{Landmark positions.} Do these clusters reflect landmark positions? Fig.~\ref{fig:clusters} (bottom) already hints at the answer: clusters of features extracted from $FC_5$ appear to contain aligned images, implying that this is indeed the case.

\begin{figure}[t!]
\begin{center}
\includegraphics[width=.9\linewidth,clip,trim = 5mm 3mm 15mm 1mm]{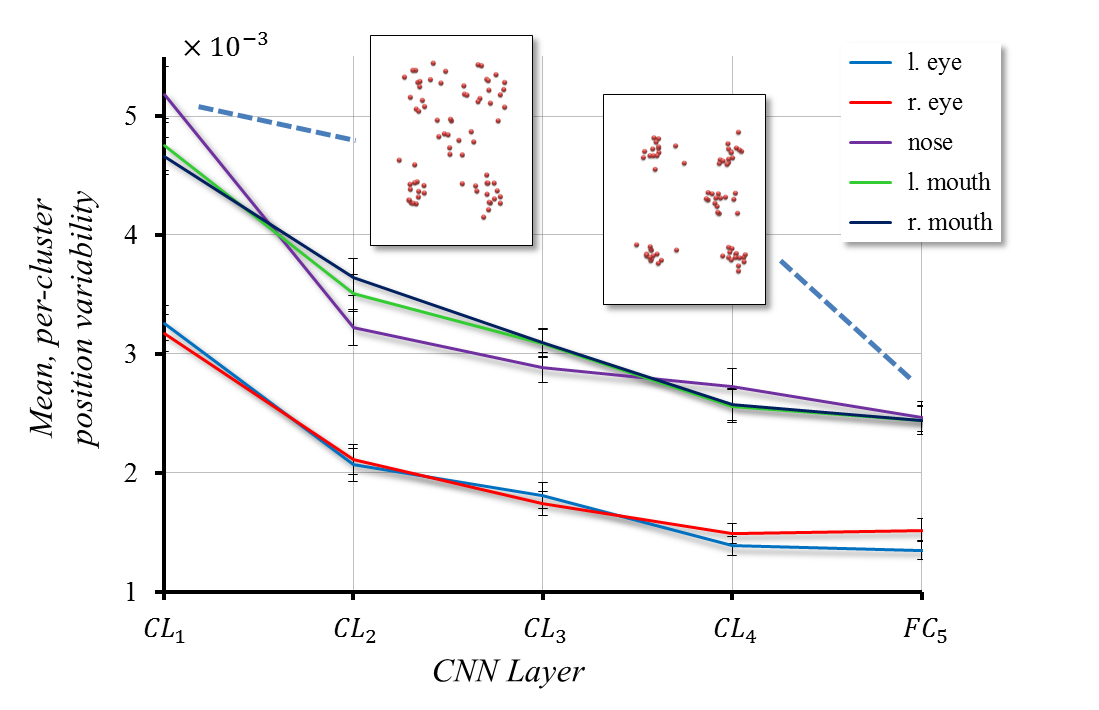}
\end{center}
\caption{{\em Landmark position variability at different layers.} $\lambda_{l,k,j}$ averaged ($\pm$ SE) over $K$ clusters in each layer (Eq.~\ref{eq:meanlandmarks}). Scatter plots for the ground truth landmarks of 15 faces from the most variable clusters of RGB (left) and $FC_5$ (right) layer features are shown, visualizing the reduced variability at the deeper layer.} \label{fig:variabilityPos}
\vspace{-3mm}
\end{figure}

We analyze this empirically using the training set images, $I_i$, and their ground truth landmarks $\hat{\mathbf{p}}_{i,j\in 1\hdots m}=(x_{i,j},y_{i,j})$. For each layer $L_{l}$ and each cluster $C_{l,k}$, we measure the variance $\lambda_{l,k,j}$ along the principle axis of each set of 2D points, $\{\hat{\mathbf{p}}_{i,j} | {\mathbf f}_{i,l} \in C_{l,k}\}_{l,k,j}$. These are then averaged for all clusters in each layer:
\begin{equation}
\mu^P_{l,j} = \frac{1}{K}\sum^{K}_{k=1}{\lambda_{l,k,j}}.\label{eq:meanlandmarks}
\end{equation}
Fig.~\ref{fig:variabilityPos} reports these values for the five landmarks, over the layers $L_l$, along with their standard errors. The average intra-cluster location variances drop by half from the input to $FC_5$. This is remarkable, as it can be interpreted to suggest that the network naturally performs hierarchical, coarse to fine feature localization, where deeper layers better represent landmark positions.

\minisection{Facial attributes.} Reexamining Fig.~\ref{fig:clusters} (bottom), it is also apparent that by aggregating similarly aligned images, clusters capture more than just head pose: some cluster centers clearly show expressions and gender. This is not surprising, as landmark positions and facial attributes are, of course, very much related. In the past, this connection was mainly utilized going from landmark positions to attribute prediction (some recent examples including~\cite{cao2014displaced,demirkus2015hierarchical,ding2013facial,LH:CVPRw15:age}). To our knowledge, the reverse direction of using attributes to predict landmarks, was only recently proposed by~\cite{zhang2014facial,zhang2015learning}.


We explore the relation between clusters and facial attributes using the following binary attribute labels, provided for the training images by~\cite{sun2013deep,zhang2014facial}: {\em male/female}, {\em smiling/Not-smiling} and {\em yes/no wearing eyeglasses}. The variance of $1/0$ values for an attribute in cluster $C_{l,k}$ is denoted by $\sigma^2_{l,k,a}$ ($a$ indexing the three attributes). A low value of $\sigma^2_{l,k,a}$ reflects a label's uniformity in a cluster. We average these variances over all clusters in each layer:

\begin{equation}
\mu^A_{l,a} = \frac{1}{K}\sum^{K}_{k=1}{\sigma^2_{l,k,a}}.\label{eq:meanattributes}
\end{equation}

\begin{figure}[t!]
\begin{center}
\includegraphics[width=.80\linewidth,clip,trim = 1mm 3mm 1mm 1mm]{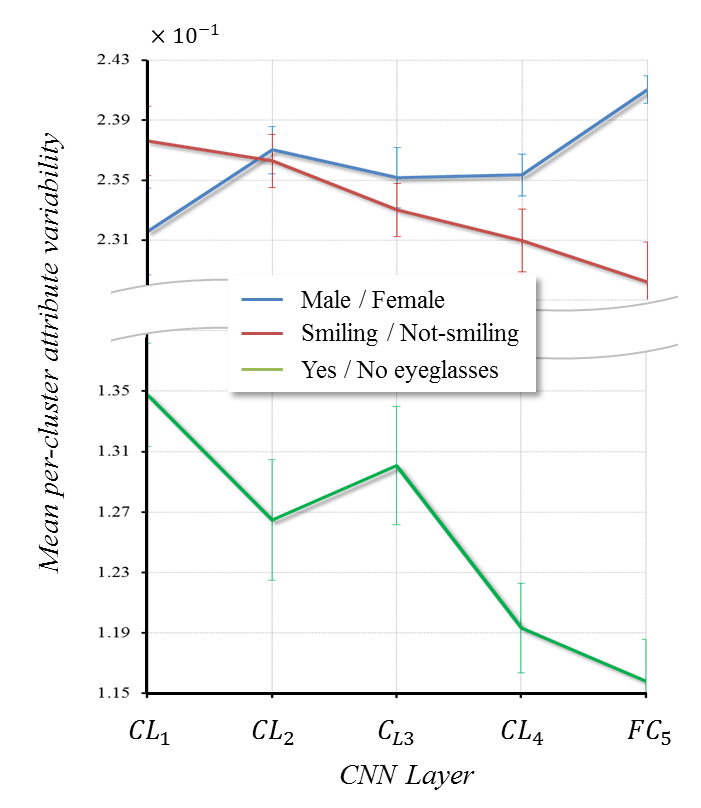}
\end{center}
\caption{{\em Variance of attribute labels at different layers.} $\sigma^2_{l,k,a}$ averaged ($\pm$ SE) over $K$ clusters in each layer (Eq.~\ref{eq:meanattributes}).} \label{fig:variabilityAttrib}
\vspace{-3mm}
\end{figure}

These values are reported in Fig.~\ref{fig:variabilityAttrib}. Higher layer clusters appear far less variable in smiling/not-smiling faces. Due to the small number of positive eyeglass attributes in the data (15.3\%, compared with 57.2\% for smiling), this attribute is less varied to begin with, but becomes far less so in higher layer clusters. These results are expected: landmark positions and appearances are heavily influenced by these two attributes. Gender, however, seems to become more varied in higher layers. This is surprising, as gender was very beneficial to detection in~\cite{zhang2014facial}. A possible explanation is that in the low resolution, $40\times 40$ pixel images used here, landmark position and appearance differences between genders are not substantial enough to aid the CNN in detection. Finally, these results also indicate that clustering images based solely on pose, without considering other facial attributes, as in e.g.,~\cite{yang2015face}, may not be optimal.

\subsection{Discussion: What can we learn from all this?}\label{sec:findings}
These findings show that when trained to regress facial landmark positions, a network internally learns representations which discriminate between differently aligned faces. In particular, when reaching the input to the first fully connected layer (represented by a red band in Fig~\ref{fig:system} (left)) {\em we already know rough head pose}. With this information we can train {\em pose specific} landmark regressors. In the past, others proposed similar two-step approaches: beginning with rough localization and then using it to accurately detect landmarks  (e.g.,~\cite{yang2015face,liang2015unconstrained,zhang2014coarse,zhang2014facial,zhu2015face,yang2015mirror}). Compared to these and others, however, here, rough pose {\em emerges naturally} from the network. As we next show, so do the final landmark regressors.

\section{The Tweaked CNN model}\label{sec:tweak}
Our analysis in Sec.~\ref{sec:analyze} is consistent with previous reports on the specialization of late layers in deep networks (e.g., the recent work of~\cite{Aubry15,yosinski2014transferable}). To our knowledge, however, the nature of this effect was never previously explored for face images or exploited to improve performance.

We propose to utilize this to improve accuracy by fine-tuning multiple versions of the final network layers: each one trained using only images represented by similar mid-network features. Because similar features imply similar landmark positions (Sec.~\ref{sec:analyze}), this can be considered teaching the network to specialize in specific facial poses and expressions. In order to avoid overfitting and improve the effectiveness of this tweaking, Sec.~\ref{sec:augment} further presents an alignment-sensitive data augmentation method.

\subsection{Tweaking by fine-tuning}\label{sec:finetune}
Our TCNN is illustrated in Fig.~\ref{fig:system} (right). We begin by training the vanilla CNN for facial landmark regression (Sec.~\ref{sec:vanilla}). Once trained, it is used to extract features from the input to $FC_5$ and aggregate them into $K$ GMMs using EM (Sec.~\ref{sec:analyze}). Next, we fine-tune the remaining weights, from $FC_5$ to the output, separately for each cluster using only its images. Here {\em early stopping} is used to fine-tune each sub-network; that is, if validation loss does not improve for 50 epochs, we cease fine-tuning that cluster. 

We emphasize that fine-tuning only involves the weights in final layers; previous layers are kept frozen. In addition, fine-tuning the network for each cluster uses the same layer dimensions (weight arrangements), commencing from the vanilla network weights. Thus, TCNN training requires little effort beyond training the initial vanilla CNN. This should be compared with methods that train multiple CNN models for multiple resolutions~\cite{zhang2014coarse} or parts (e.g.,~\cite{liang2015unconstrained}).

In practice, training the vanilla CNN on a dedicated machine required $\sim$6.5 hours. By comparison, tweaking a single cluster took $\sim$2 {\em minutes}. Of course, tweaking different clusters can be performed on separate machines in parallel. Consequently, the added time for tweaking is very low.

Estimating landmark positions for a query photo $I_Q$ begins by using the vanilla CNN to extract its $FC_5$ feature, ${\mathbf f}_{Q,5}=f(I_Q,FC_5)$. It is then assigned to the cluster $C_{5,k}$ which maximizes the posterior probability (Eq.~\ref{eq:gmm}). From this point onward, the network proceeds processing this feature vector using only the layers fine-tuned for $C_{5,k}$, finally returning the output from $FC^k_6$.

This process treats horizontally flipped versions of the same face differently -- by different tweaked processes -- and one may be better than the other. To address this, we evaluate test images twice, also processing images after horizontal mirroring. Each two predictions are then averaged (after mirroring the coordinates of the flipped image) to obtain final results. 

\minisection{Why $FC_5$?} Fig.~\ref{fig:variabilityPos} suggests that input features to $CL_4$ are as capable as $FC_5$ in representing different alignments. In addition, its position further from the output leaves two layers to fine-tune, potentially allowing for better specialization to the appearances in each cluster. We tested TCNNs with both layers and found that the performance gained by using $CL_4$ is comparable to using $FC_5$, yet requires more time to tweak. Our design therefore uses $FC_5$.


\subsection{Alignment-sensitive data augmentation}\label{sec:augment}
The numbers of training images in each cluster, reported in Table~\ref{tab:clusterstats}, are clearly nowhere near enough to train even the last layers of the network without risking overfitting. One means of overcoming this is by augmenting the training data. Popular methods for doing so include {\em oversampling}~\cite{krizhevsky2012imagenet} -- essentially producing multiple, slightly translated versions of the input image by cropping it at different offsets -- and {\em mirroring} the images~\cite{yang2015mirror}.

Applying these here, however, proved unsuccessful. This is not entirely surprising: Each tweaked, fine-tuned network trains on representations from the same cluster. These, as we showed earlier, should all be well aligned. Oversampling and mirroring both introduce misaligned images into each cluster, increase landmark position variability and so undermine the goal of our fine-tuning.

Ostensibly, we could add data to each cluster by sampling from the GMM components (see, Eq.~\ref{eq:gmm}). These, however, are defined on intermediate features and cannot produce landmark coordinates required for training.

Instead, we propose augmenting the image set used to fine-tune tweaked layers in an {\em alignment-sensitive} manner. Let $I_i, I_{i'}$ be two training images, randomly selected as being associated to the same cluster, $C_{5,k}$, and $\hat{\mathbf{p}}_{i,j}, \hat{\mathbf{p}}_{i',j}, j\in 1\hdots,m$ their $m$ respective ground truth landmarks. We estimate the non-reflective similarity transform, $\mathbf{H}$ mapping $\hat{\mathbf{p}}_{i',j}$ to $\hat{\mathbf{p}}_{i,j}$ using standard least squares~\cite{hartley2003multiple} and use it to backwards warp $I_i$ to the coordinate frame of $I_{i'}$, i.e., 
\begin{equation}
{I'_i(x,y)}\triangleq I_i(\mathbf{H}^{-1}(x,y)).\label{eq:warp}
\end{equation} 
\noindent The new image, $I'_i$ is verified to belong to $C_{5,k}$ by extracting its feature representation from the input to $FC_5$ and associating it with a cluster $C_{5,k'}$ using Eq.~\ref{eq:gmm}. If $k\neq k'$, then the generated image does not belong in the same cluster with the two images used to produce it and is therefore rejected. In practice, $<$40\% of the generated images failed this test, typically due to artifacts introduced by warping. This should be compared with over 96\% rejected when using images from other clusters. Accepted images $I'_i$ are added to the training along with the landmark labels of~$I_{i'}$.

\begin{figure}[t!]
\begin{center}
\includegraphics[width=.95\linewidth,clip,trim = 3mm 3mm 3mm 1mm]{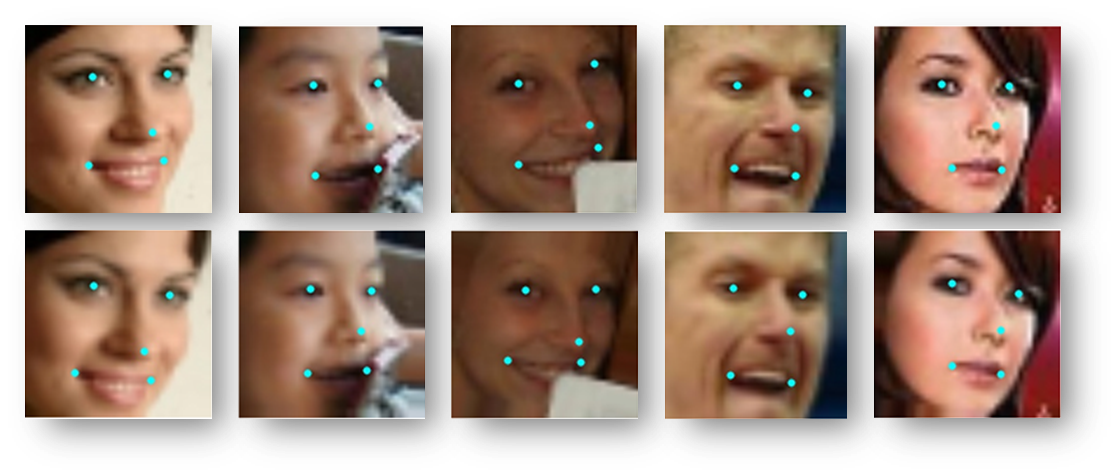}
\end{center}
\vspace{-2mm}
\caption{{\em Alignment-sensitive data augmentation.} Top: Training images from the same $FC_5$ cluster. Bottom: Images added to the cluster to increase training set size. Bottom images are noticeably different from their origins in the top, yet remain in their original cluster. Ground truth landmark positions used to align these images appear in cyan.} \label{fig:augmented}
\vspace{-3mm}
\end{figure}

This data augmentation approach can presumably be used with any number of clusters, particularly, when there is only one: to augment the data used to train the vanilla CNN. In such cases, however, the rejection step mentioned above is meaningless and all generated images are used for training. In practice, training the CNN with this data augmentation technique failed to show improved results and so we do not apply it for the single cluster, vanilla CNN results reported in Sec.~\ref{sec:results}.

Fig.~\ref{fig:augmented} provides a few examples of images added by this process to our training. These are slightly, but noticeably misaligned with their sources. They therefore introduce variation to each tweaking training set, yet still belong to their original clusters. We use this process to artificially raise the number of training images in each cluster to $5,000$ images. We empirically found smaller numbers of augmented images to result in overfitting and larger numbers to provide no meaningful performance gain.

\section{Comparison with existing work}\label{sec:previouscompare}
We next compare our TCNN design with relevant existing models and detectors. 

Several previous methods proposed fine-tuning late layers to specific data (see~\cite{yosinski2014transferable,oquab2014learning,tzeng2015simultaneous} for recent examples). These, however, focus on domain transfer applications where fine-tuned layers are trained with new supervised data from different problem domains. We, by contrast, improve network performance on augmented subsets of the original data, determined in an unsupervised manner. 


Our alignment sensitive method for data augmentation (Sec.~\ref{sec:augment}) may be viewed as a particular instance of sampling from CNN feature space (e.g.,~\cite{gregor2015draw}). Our approach is very different from those methods. Furthermore, they did not use their sampling to generate additional training data.


TCNN can be considered to operate in a multi-scale manner, where coarse landmark positions are reflected by intermediate representations and then localized by the tweaked, final layers. Multi-scale network designs were proposed in the past and we refer to~\cite{Xie2015Holistically} for a very recent survey. None of these, however, uses internal representations to guide multi-scale processing as we do. As we mention in Sec.~\ref{sec:findings}, though some landmark detection methods also take a course to fine approach, these are very different from the one proposed here.


\begin{figure}[t!]
\begin{center}
\includegraphics[width=.8\linewidth,clip,trim = 3mm 3mm 3mm 1mm]{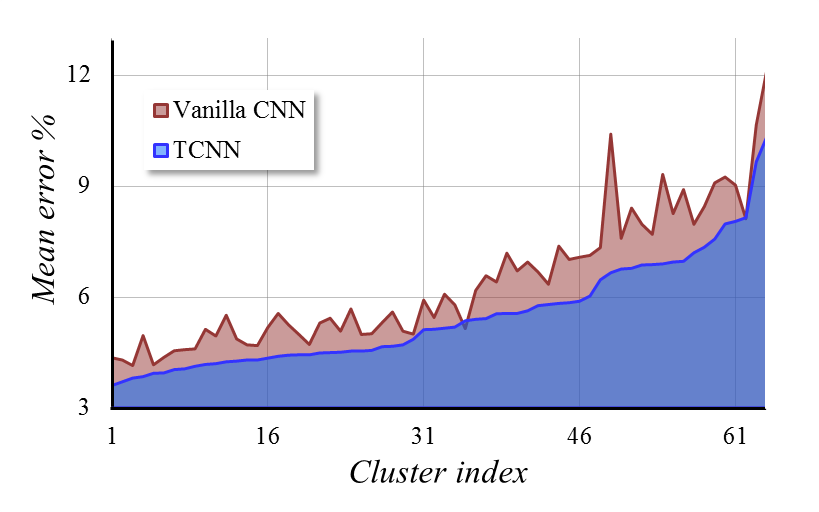}
\end{center}
\vspace{-2mm}
\caption{{\em Effect of tweaking.} Validation set, per cluster mean error rate for vanilla CNN (in red) vs. TCNN (blue) (Sec.~\ref{sec:tweak}). $K=64$ clusters were produced from $FC_5$ features, sorted here by TCNN performance. Lower values are better.} \label{fig:analysis:finetune}
\vspace{-3mm}
\end{figure}

\begin{figure}[t!]
\begin{center}
\includegraphics[width=.8\linewidth,clip,trim = 5mm 3mm 6mm 15mm]{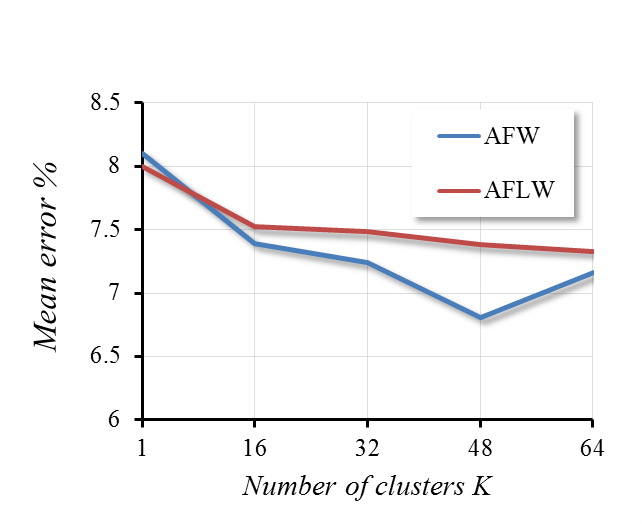}
\end{center}
\vspace{-2mm}
\caption{{\em Number of clusters vs. mean error.} Results on AFW~\cite{zhu2012face} (blue) and AFLW~\cite{kostinger2011annotated} (red) for TCNN models with different numbers of tweaked processes, $K$. Lower numbers are better.} \label{fig:analysis:clusters} 
\vspace{-3mm}
\end{figure}

Finally, the recent work of~\cite{zhang2014facial,zhang2015learning} detects facial landmarks with a network trained using multi-task learning of both landmarks and attributes. To do so, they manually specify facial attributes and label the training images accordingly. We use no such auxiliary labels, we show in Sec.~\ref{sec:analyze} that the network naturally learns to discriminate between some of these attributes, and next show our TCNN to exceed their performance, with a simpler network design. 

\section{Experiments}\label{sec:results}
\subsection{Landmark detection}\label{sec:results:detect}
\minisection{Evaluation criteria.} In our tests we use the detector {\em error rate} to report accuracy (see~\cite{dantone2012real} and many others since). It is computed by normalizing the mean distance between predicted to ground truth landmark locations to a percent of the inter-ocular distance. 

\begin{figure*}[t!]
\centering{
\includegraphics[width=.95\linewidth,clip,trim = 3mm 3mm 3mm 5mm]{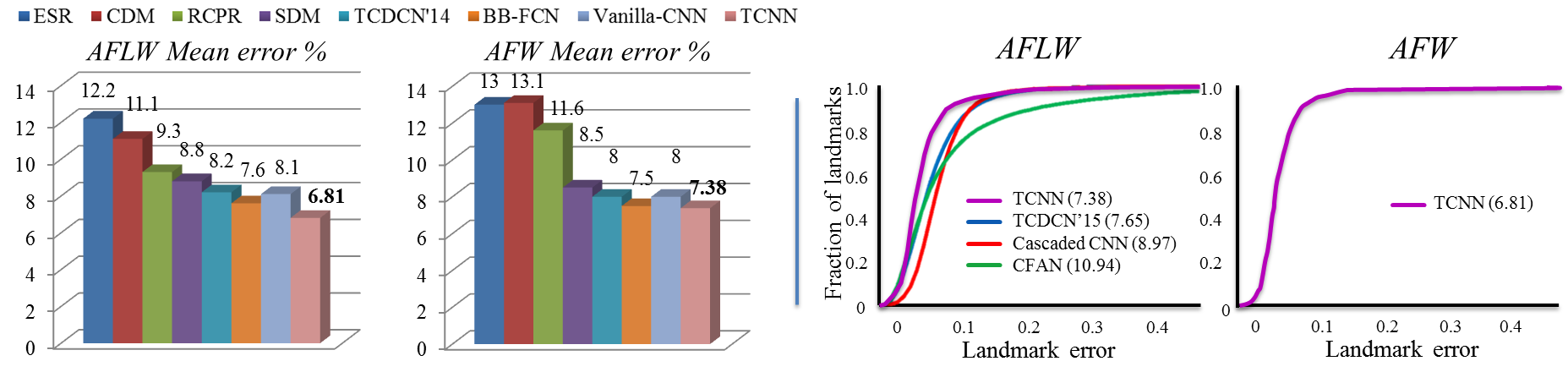}
}
\caption{{\em AFLW and AFW results.} Left: Error rates on the AFLW~\cite{kostinger2011annotated} and AFW~\cite{zhu2012face} benchmarks for ESR~\cite{cao2014face}, CDM~\cite{yu2013pose}, RCPR~\cite{burgos2013robust}, SDM~\cite{xiong2013supervised}, TCDCN'14~\cite{zhang2014facial} and BB-FCN~\cite{liang2015unconstrained}, vs. our vanilla CNN (Sec.~\ref{sec:vanilla}) and TNCC. Lower values are better. Right: Accumulative error curves reported for CFAN~\cite{zhang2014coarse}, Cascaded CNN~\cite{sun2013deep} and TCDCN'15~\cite{zhang2015learning} vs. our TCNN, along with mean error \%.} \label{fig:empirical}
\end{figure*}

\begin{figure*}[t!]
	\centering{
	\begin{tabular}{cc@{~~~}cc}
    \rotatebox{90}{\hspace{07mm}{\em AFLW}} &
        \includegraphics[width=.43\linewidth,clip,trim = 3mm 3mm 3mm 1mm]{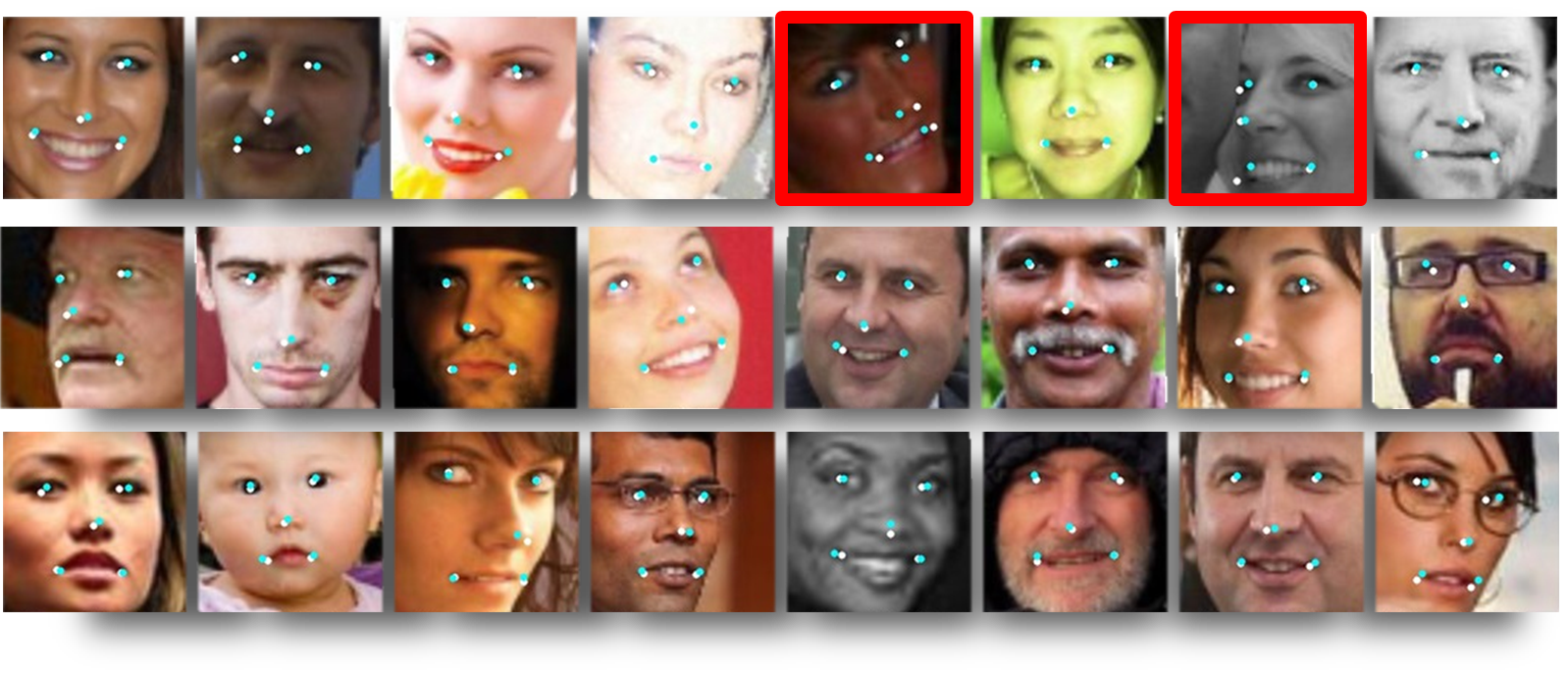} &
        \rotatebox{90}{\hspace{08mm}{\em AFW}} &
		\includegraphics[width=.43\linewidth,clip,trim = 3mm 3mm 3mm 1mm]{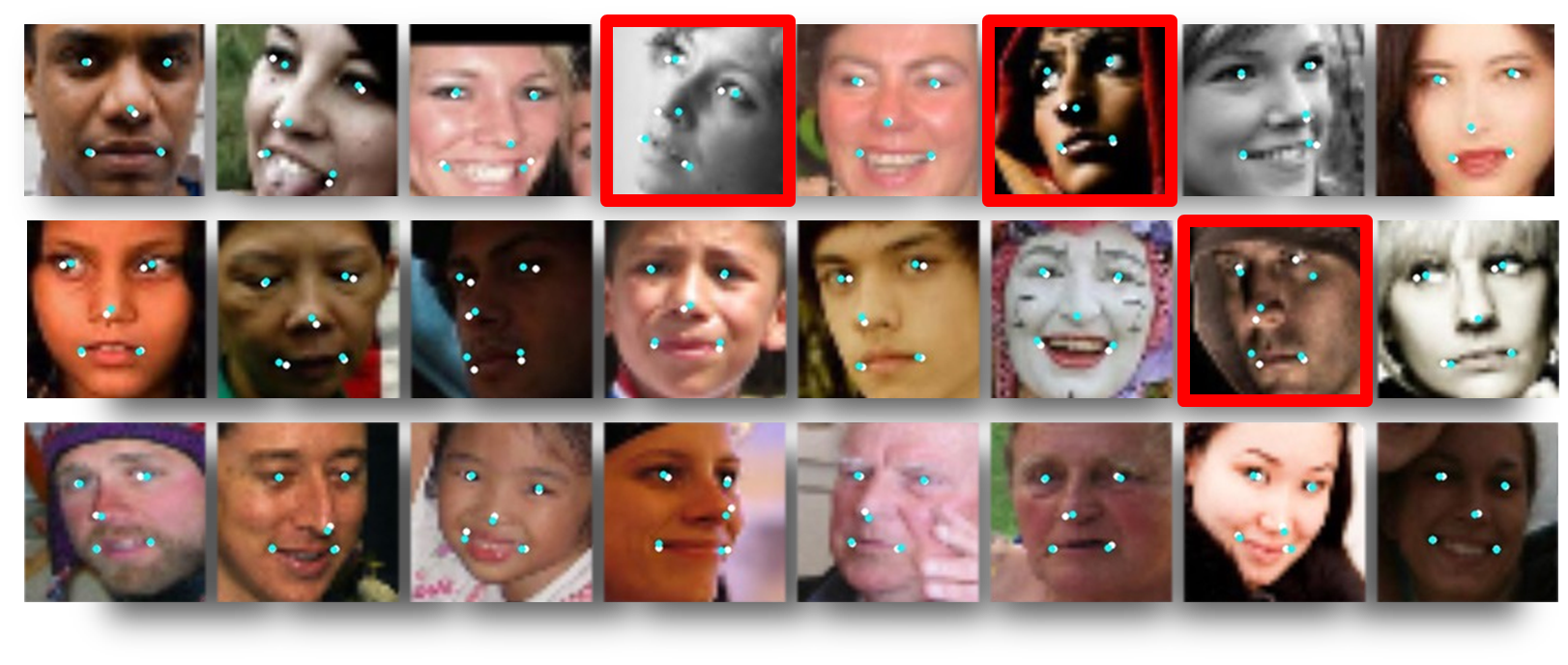}
		\vspace{-3mm}
	\end{tabular}
	}
	\caption{{\em Example detection results.} Qualitative detections on AFLW~\cite{kostinger2011annotated}  (left) and AFW~\cite{zhu2012face} (right). Showing ground truth landmark locations in white and our TCNN detections in cyan. Typical mistakes highlighted in red.}
	\label{fig:qualitative}
\end{figure*}

\minisection{The effect of tweaking by fine-tuning.} Fig.~\ref{fig:analysis:finetune} demonstrates the effect of tweaking by fine-tuning (Sec,~\ref{sec:finetune}) with alignment-sensitive data augmentation (Sec.~\ref{sec:augment}) on landmark detection accuracy. It reports the per-cluster detection error rate on validation set images, comparing vanilla CNN performance with TCNN. For convenience, cluster labels are sorted by ascending TCNN errors. Apparently, in nearly all clusters, TCNN manages to improve accuracy, in some cases, by several percents. 

We note that the cluster centers presented in Fig.~\ref{fig:clusters} (bottom) are also sorted (left-to-right, top-to-bottom) by the TCNN errors of Fig.~\ref{fig:analysis:finetune}. Evidently, the worst performing clusters are those where poses were non-frontal, with the worst performing cluster in Fig.~\ref{fig:clusters} (bottom) containing near profile views. These poses are underrepresented in the training set and so it is not surprising that detection accuracy is lower in those clusters. Importantly, these are also the clusters where TCNN was most effective, providing the biggest performance improvements.


\minisection{Benchmarks.} We evaluate our TCNN on two standard benchmarks for landmark detection: the Annotated Face in-the-Wild (AFW) of~\cite{zhu2012face}, containing 468 faces, and the Annotated Facial Landmarks in the Wild (AFLW)~\cite{kostinger2011annotated} with its 24,386 faces, using the same test subsets from~\cite{zhang2014facial}. Both sets include faces from Flickr albums, manually annotating with five facial landmarks. Both therefore represent unconstrained settings with many of the appearance variations our method is expected to handle.


\minisection{Effect of $K$.} We evaluate how the number of clusters, $K$, affects overall accuracy. Fig.~\ref{fig:analysis:clusters} reports error rates for both the AFW and AFLW benchmarks with varying cluster numbers. Using more tweaked final layers appears to improve performance. This, however, is only up to a point: Splitting the training data into too many clusters produces clusters which do not have enough examples for effective fine-tuning. In fact, beyond $K=64$ clusters, fine-tuning the final network layers often resulted in overfitting. The results reported next use $48$ clusters for our TCNN implementation.

Also interesting is the tweaking effort (not shown). Measured in epoch numbers, the effort ($\pm$ SD) required to fine-tune each tweaked process for the different values of $K$ is 79.63 $\pm$ 2.9. Hence, this effort grows linearly with $K$.

\begin{table*}[!t]
\footnotesize{
\caption{{\em Janus CS2 verification.} Comparing performance obtained on faces aligned with four landmark detectors. Higher values are better.}\label{tab:janus}
\vspace{-2mm}
\begin{center}
\begin{tabular}{l|ccc|cccc}
\toprule
 & \multicolumn{3}{|c|}{Baseline / reference methods} & \multicolumn{4}{c}{Same method, different alignments} \\
TAR $@$ & COTS1~\cite{klare2015pushing} & FV~\cite{Simonyan13} & DCNN~\cite{chen2015unconstrained} & CLNF~\cite{baltrusaitis2013constrained} & ERT~\cite{kazemi2014one} & LBF~\cite{ren2014face} & TCNN \\ \hline
FAR=1e-2 & .581 $\pm$ .05 & .411 $\pm$ .08 & .649 $\pm$ .01 & .677 $\pm$ .01 & .689 $\pm$ .01 & .643 $\pm$ .02 & {\bf .718 $\pm$ .01} \\  
FAR=1e-1 & .767 $\pm$ .01 & .704 $\pm$ .02 & .855 $\pm$ .01 & .897 $\pm$ .00 & .894 $\pm$ .00 & .888 $\pm$ .00 & {\bf .903 $\pm$ .00} \\  
\bottomrule
\end{tabular}
\end{center}
}
\vspace{-9mm}
\end{table*}

\minisection{Comparison.} The following facial landmark detectors are compared with our approach: ESR~\cite{cao2014face}, CDM~\cite{yu2013pose}, RCPR~\cite{burgos2013robust}, SDM~\cite{xiong2013supervised}, TCDCN'14~\cite{zhang2014facial}\footnote{Differrent results were reported for TCDN in the original conference paper from 2014~\cite{zhang2014facial} and its journal version from 2015~\cite{zhang2015learning}. We compare with both, and, when relevant, denote the year in the method's name.} and BB-FCN~\cite{liang2015unconstrained}. Fig.~\ref{fig:empirical} (left) shows our results vs. previously reported performances from~\cite{liang2015unconstrained}, on the same benchmark protocols. The only exception, BB-FCN, was trained on twice the images. Fig.~\ref{fig:empirical} (right) provides also accumulative error curves for methods which reported this information. Lastly, Fig.~\ref{fig:qualitative} illustrates some TCNN detections. 

The vanilla CNN is on-par with the state of the art. TCNN improves this, surpassing all other results in both benchmarks, cutting down {\bf 8\%} of the error on AFLW and~{\bf 17\%} on AFW. To emphasize the significance of this, we note that the performance gaps between us and the best published result~\cite{zhang2014facial} is 1.4\% on AFW and 0.62\% on AFLW, whereas their reported gaps were 0.6\% and 0.5\%,~resp. 

Furthermore, the combined benefit of more data and larger networks was previously noted by, e.g.,~\cite{simonyan2014very}, and can at least partially explain the performance of BB-FCN~\cite{liang2015unconstrained}. Our results show that {\em better performance} can be achieved by careful processing using a {\em simpler model} and {\em less data}.


\subsection{Face recognition}\label{sec:results:recog}
A popular reason for detecting facial landmarks is alignment for face recognition (e.g.,~\cite{hassner2015effective}). We therefore also provide face verification results using faces aligned by different landmark detection methods. This also allows for comparisons of methods trained to detect different features or feature numbers.

\minisection{The Janus Benchmark.} We use the Janus CS2 face verification benchmark~\cite{klare2015pushing}. This benchmark is selected here not only because it is newer (and hence less saturated) than other benchmarks, but also due to the high pose variations exhibited by its faces: as part of its design methodology, images were collected for this set with an emphasis on increasing pose range, including near profile views, almost nonexistent in previous benchmarks (e.g, LFW~\cite{LFWTech}).

The Janus collection includes $5,712$ images and $2,042$ videos (represented by $20,412$ frames) of 500 subjects. Subjects are represented not by single images, but by {\em templates} containing one or more images. The CS2 protocol provides a training set with 333 subject templates and ten test splits of template pairs. The objective is to determine if two test templates belong to the same person (i.e., {\em same}/{\em not-same} classification). Verification accuracy is measured using the true accept rate (TAR) at false accept rates (FAR) $1e-1$ and $1e-2$.

\minisection{Recognition pipeline.} To focus on the contribution of our TCNN detector, we use the same recognition pipeline, comparing verification accuracy on images aligned with TCNN landmarks vs. CLNF~\cite{baltrusaitis2013constrained}, ERT~\cite{kazemi2014one} and LBF~\cite{ren2014face}. These were selected due to their publicly available implementations and to complement the ones in Sec.~\ref{sec:results:detect}. In all cases, we use the face bounding boxes from~\cite{king2009dlib}. Detected features were then used for non-reflective similarity transform alignment~\cite{hartley2003multiple}.

Aligned images are processed using the standard AlexNet~\cite{krizhevsky2012imagenet}, fine-tuned on CASIA WebFace images~\cite{yi2014learning}. No additional fine-tuning was performed on Janus training data. We extract the output values of the CNN's penultimate layer ($FC_8$), as raw features. These are power normalized~\cite{sanchez2013image} and projected down from 4,096D to 4,000D using PCA computed on the Janus training set images. 

Given templates $T$ and $T'$ we compute the Pearson correlation coefficients for matching PCA projected features of all image pairs $(I,I'), I\in T, I'\in T'$. A final template-to-template similarity score is obtained by the SoftMax~\cite{krizhevsky2012imagenet} of these scores.

\minisection{Results.} Performances are reported in Table~\ref{tab:janus}. As reference, we provide previously published results for the COTS1, off-the-shelf method~\cite{klare2015pushing}, as well as FV~\cite{Simonyan13} and DCNN~\cite{chen2015unconstrained} (without fine-tuning on Janus training data, not performed by us either). Apparently, all other system components being equal, images aligned using TCNN are easier to recognize than those of other detectors. In fact,  by using TCNN aligned images, our recognition pipeline outperforms the larger, dedicated DCNN network, along with its landmark detector of~\cite{asthana2014incremental}.


\begin{figure*}[t]
\centering
\subfloat[49 pnt. accum. errors]{
\includegraphics[width=.3\textwidth,clip,trim = 2mm 2mm 0mm 10mm]{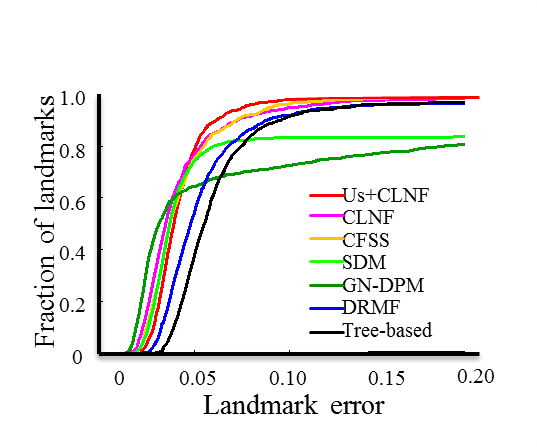}
\label{fig:300w:errorcurves49}
}
\subfloat[68 pnt. accum. errors]{
\includegraphics[width=.3\textwidth,clip,trim = 2mm 2mm 0mm 10mm]{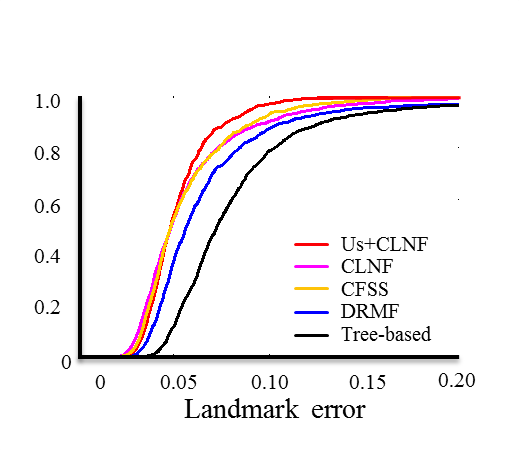}%
\label{fig:300w:errorcurves68}
}
\subfloat[Performance]{
\raisebox{.23\height}{\scriptsize{
\begin{tabular}[b]{lcccc}
\toprule
& \multicolumn{2}{c}{49 Points} & \multicolumn{2}{c}{68 Points}\\
\textbf{Method}& \textbf{@10\%}& \textbf{MER} & \textbf{@10\%}& \textbf{MER}\\ \hline
Tree-based$^1$ & 8.48 &.0626 & 21.25 & .0828\\
DRMF$^1$ & 8.09 &.0564 & 12.96 & .0675\\
GN-DPM &27.39 &.1141 & -- & -- \\
SDM$^1$ & 16.86 &.0411 & -- & -- \\
CLNF & 5.17 &.0483 & 10.14 & .0630\\ 
CFSS$^2$ & 3.77 & .0460 & 7.26 & .0586\\ \hline 
Us+CLNF & {\bf 1.74} &.0433 & {\bf 3.49} & .0537\\
\bottomrule
\end{tabular}
}
}
\label{fig:300w:errors}
}
\vspace{-2mm}
\caption{{\em 300W results.} (a) Accumulative error curves for 49 point detections; (b) 68 point detections; (c) \% images with 49 (68) landmark detection errors higher than \%10 inter-occular distances and mean error rates (MER) for 49 and 68 point detections. Lower numbers are better. $^1$~Mean error rates for true positive detected faces; $^2$ Not including~AFW.}
\label{fig:300w}
\vspace{-4mm}
\end{figure*}

\begin{figure*}[t!]
	\centering{
	\begin{tabular}{c@{~~~~~}c}
        \includegraphics[width=.48\linewidth,clip,trim = 3mm 3mm 3mm 1mm]{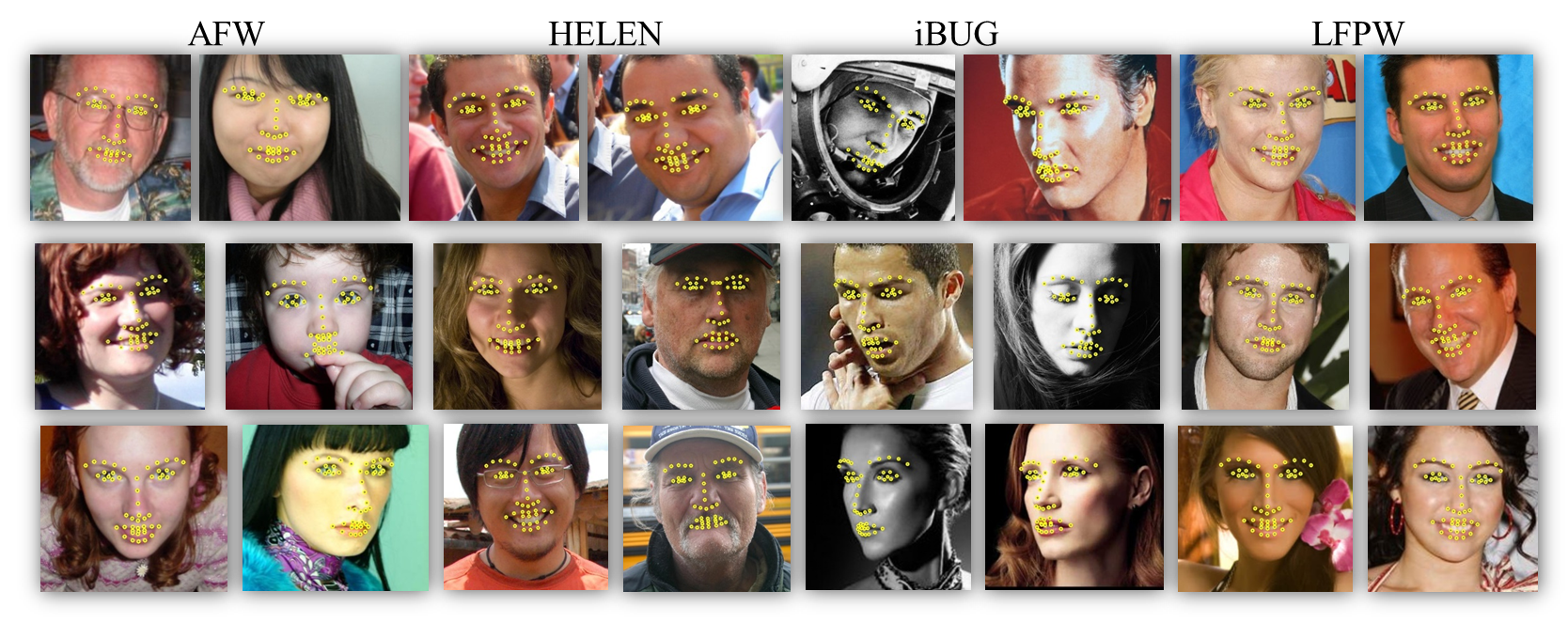} &
		\includegraphics[width=.48\linewidth,clip,trim = 3mm 3mm 3mm 1mm]{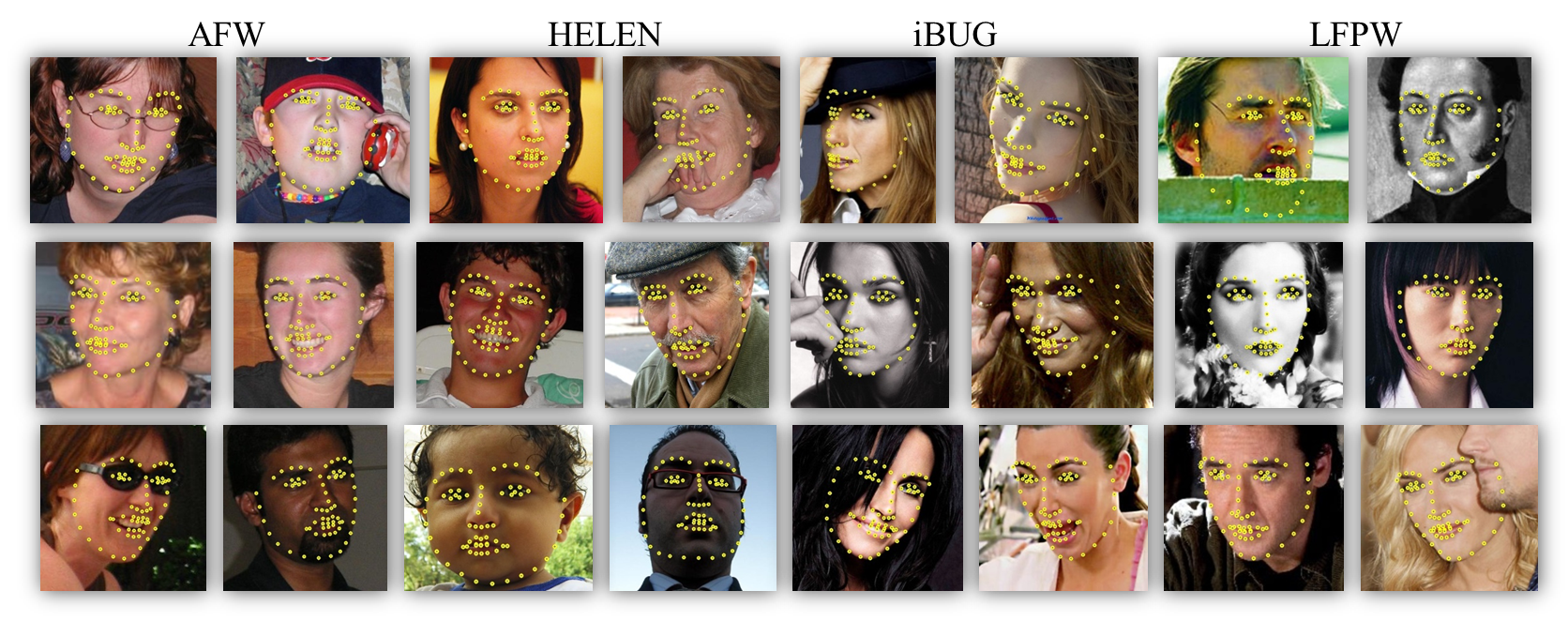}
		\vspace{-3mm}
	\end{tabular}
	}
	\caption{{\em Example detection results on the 300W benchmark.} Qualitative detections of our system for 49 landmarks (left) and 68 landmarks (right). Each pair of columns presents images from one of the four subsets of the 300W benchmark, from left to right: AFW, HELEN, iBUG and LFPW.}
	\label{fig:300w:qualitative}
	\vspace{-6mm}
\end{figure*}

\subsection{Beyond five landmarks}\label{sec:68}
Our results in Sec.~\ref{sec:results:recog} suggest that to align faces for face recognition, five points may be enough (or better) than the larger landmark sets detected by other methods. For some applications, however, more landmarks may be necessary. Though our system is not specifically tailored to particular landmark numbers or arrangements, rather than retraining it for these tasks, we instead simply use its five landmarks to initialize an existing, publicly available 49 or 68 point landmark detector, here CLNF of~\cite{baltruvsaitis2014continuous}.

We evaluate performance on the 300W data set~\cite{sagonas2015300}. It is the largest, most challenging benchmark of its kind, containing images from the LFPW~\cite{belhumeur2013localizing}, HELEN~\cite{le2012interactive}, AFW~\cite{zhu2012face} and iBUG~\cite{sagonas2013300} collections. Since we did not use AFW for training our method, as others have done in the past, we can use the entire 300W collection for testing. Once again we use the face detector from~\cite{king2009dlib} to find face bounding boxes. Whenever it failed to detect a face, we defaulted to the ground truth bounding boxes provided in 300W and used by the other baselines. The five points detected by our method are then used to initialize the response maps for the corresponding local patches in~\cite{baltruvsaitis2014continuous}.

We compare our approach (Us+CLNF) to the Tree-based method of~\cite{zhu2012face},  DRMF~\cite{asthana2013robust}, GN-DPM~\cite{tzimiropoulos2014gauss}, SDM~\cite{xiong2013supervised}, CFSS~\cite{zhu2015face} and CLNF with its original initialization~\cite{baltruvsaitis2014continuous}. Note that previous reports for the performance of the Tree-based method, DRMF and SDM reflect accuracy only on true positive detected faces (generally considered easier). Moreover, CFSS was not tested on the AFW subset of 300W. Tree-based, DRMF and SDM results were taken from~\cite{baltruvsaitis2014continuous}, reporting accumulative error curves for both 49 and 68 points. Also, GN-DPM and SDM provide only 49 point detections. Fig.~\ref{fig:300w} reports these numbers. Evidently, the better initialization offered by our method allows accurately localizing landmarks for a larger number of faces.
\vspace{-3mm}
\section{Conclusions}

\begin{figure}[b!]
\begin{center}
\begin{tabular}{cc}
\rotatebox{90}{\hspace{12mm}{\em Deep intermediate FR features}} & \includegraphics[width=.75\linewidth,clip,trim = 3mm 3mm 3mm 3mm]{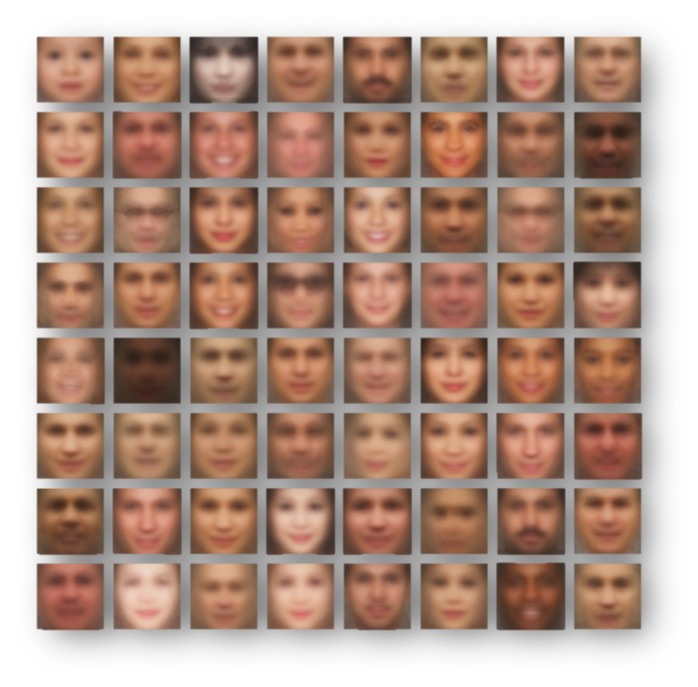}
\end{tabular}
\vspace{-3mm}
\end{center}
\caption{{\em Average images for 64 face clusters.} Clusters computed using features from an intermediate layer of a network trained for face recognition (Sec.~\ref{sec:results:recog}). Comparing these with Fig.~\ref{fig:clusters} demonstrates the far greater emphasis in these clusters on identity related features. (Note: Produced using the same images as Fig.~\ref{fig:clusters}.)} \label{fig:fr_clusters}
\vspace{-3mm}
\end{figure}

The pursuit of better landmark detection accuracy led many to propose progressively more elaborate models and representations, and use increasing amounts of data to train them. Contrary to these, we show how hierarchical, discriminative processing can naturally be introduced to an existing CNN design for facial landmark regression. This, by careful analysis and processing of the values produced at intermediate CNN layers. In so doing, we boost performance beyond those of more involved, state-of-the-art systems. 

We conclude by noting that the same analysis may conceivably be used to improve networks trained for other tasks, as an alternative to growing larger networks or using more data. Fig.~\ref{fig:fr_clusters} hints of these potentials for face recognition. It shows average faces of clusters formed from the same images of Fig.~\ref{fig:clusters}, but using $FC_7$ features extracted from our face recognition CNN (Sec.~\ref{sec:results:recog}). These clearly capture identity related attributes (e.g., facial hair, ethnicity; note ``baby cluster'' in the top left) far better than those in Fig.~\ref{fig:clusters}, suggesting that tweaking, here for identity related appearances, can be likewise effective for this task.

\section*{Acknowledgments}
This research is based upon work supported in part by the Office of the Director of National Intelligence (ODNI), Intelligence Advanced Research Projects Activity (IARPA), via IARPA 2014-14071600011. The views and conclusions contained herein are those of the authors and should not be interpreted as necessarily representing the official policies or endorsements, either expressed or implied, of ODNI, IARPA, or the U.S. Government.  The U.S. Government is authorized to reproduce and distribute reprints for Governmental purpose notwithstanding any copyright annotation thereon.

\bibliographystyle{ieee}

\end{document}